\documentclass{article}

\PassOptionsToPackage{numbers, compress}{natbib}

\usepackage[preprint]{neurips_2026}

\usepackage[utf8]{inputenc}
\usepackage[T1]{fontenc}

\usepackage{graphicx}
\usepackage{amsmath}
\usepackage{amsfonts}
\usepackage{booktabs}
\usepackage[table]{xcolor}
\usepackage{multirow}
\usepackage{pifont}
\usepackage{float}
\usepackage{caption}
\usepackage{nicefrac}
\usepackage{microtype}
\usepackage{newunicodechar}
\newunicodechar{λ}{$\lambda$}

\usepackage{url}
\usepackage{hyperref}
\title{PanoPlane: Plane-Aware Panoramic Completion for Sparse-View Indoor 3D Gaussian Splatting}

\author{
  \textbf{Adil Qureshi} \quad
  \textbf{Dongki Jung} \quad
  \textbf{Jaehoon Choi} \quad
  \textbf{Dinesh Manocha}\\
  University of Maryland, College Park\\
  \texttt{\{adilq, jdk9405, kevchoi, dmanocha\}@umd.edu}
}

\begin{document}

\maketitle


\begin{center}
  \includegraphics[width=\textwidth]{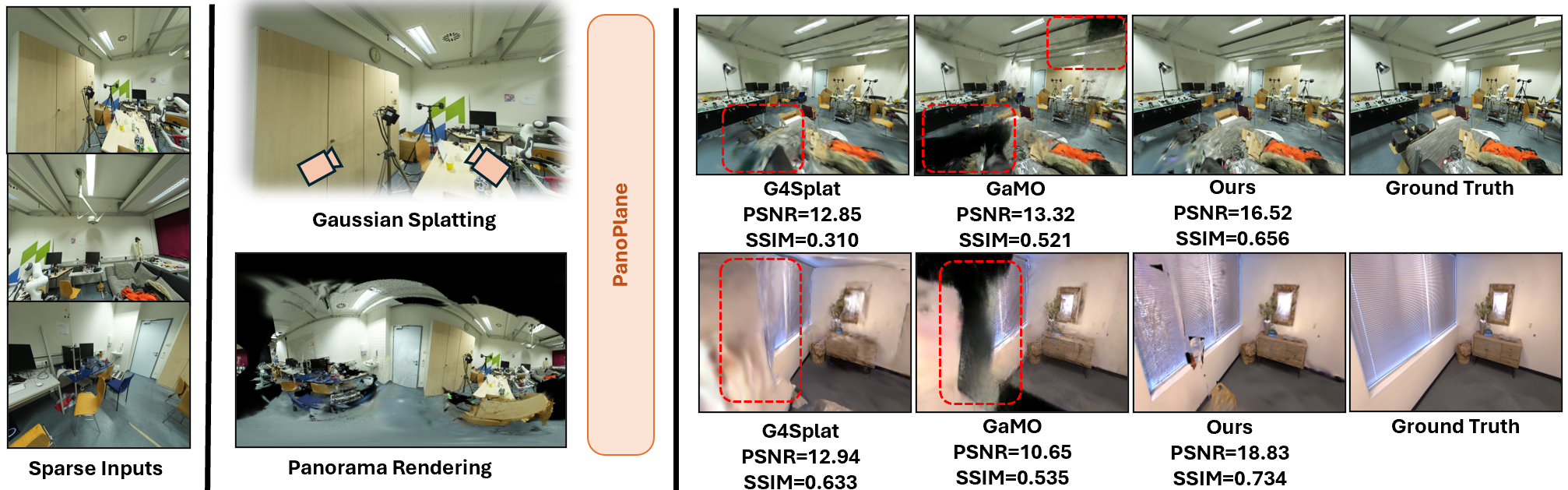}
  \vspace*{-4mm}
  \captionof{figure}{PanoPlane takes sparse input views of an indoor scene, renders a panoramic image with unobserved regions marked as holes, and completes it using layout anchored attention steering, a training-free mechanism that anchors the panoramic flow-matching model's internal attention to detected planar surfaces (walls, floors, ceilings). The completed 360° panorama provides globally consistent supervision for 3D Gaussian Splatting, achieving state-of-the-art novel view synthesis.}
  
  \label{fig:teaser}
\end{center}

\begin{abstract}
We present PanoPlane, an approach for high-fidelity sparse-view indoor novel view synthesis that reconstructs closed room geometry via panoramic scene completion. Unlike perspective-based methods that generate training views from limited fields of view, PanoPlane leverages $360^{\circ}$ panoramic completion to condition the generative process on the full spatial layout. 
We propose Layout Anchored Attention Steering, a training-free mechanism that steers attention within the diffusion model's internal representation toward scene's detected planar surfaces at inference time. By directing each unobserved region's attention toward geometrically consistent observed content, our method replaces unconstrained hallucination with grounded surface extrapolation. The resulting panoramic completions provide supervision for 3D Gaussian Splatting, enabling accurate novel-view synthesis across unobserved regions from as few as three input views. Experiments on Replica, ScanNet++, and Matterport3D demonstrate state-of-the-art novel view synthesis quality across 3, 6, and 9 input views, achieving up to $+17.8\%$ improvement in PSNR over the current state-of-the-art baseline without any training or fine-tuning of the diffusion model.

\end{abstract}

\section{Introduction}

\label{sec:intro}

Reconstructing indoor scenes from sparse viewpoints for novel view synthesis is a critical enabler for various tasks such as scene understanding, embodied robot navigation and spatial computing \cite{jin2025ral, guo2025iglnav, honda2025gsplatvnm, zheng2026dgsnav, lei2025gaussnav}.
3D Gaussian Splatting (3DGS) \cite{kerbl20233d} achieves photorealistic novel view synthesis with real-time rendering when dense multi-view coverage of the scene is available \cite{mihajlovic2024splatfields, patle2025adgs, chen2025quantifying, yin2024fewviewgs, song2026d2gs}.
However, with only a handful of input views, large portions of the scene are never observed, often leading to collapsed or no geometry, floating artifacts, and incoherent surfaces, particularly in indoor environments \cite{xiong2023sparsegs, zhu2023FSGS, li2024dngaussian, niemeyer2022regnerf, wang2023sparsenerf}.  

Recent methods address sparse-view reconstruction by leveraging diffusion priors to generate additional training views for 3DGS supervision \cite{paliwal2025ri3d, ni2026g4splat, huang2025gamo, topaloglu2026oraclegs}, employing strategies from repair-and-inpainting pipelines to video diffusion and multi-view outpainting. 
While these approaches have made significant progress, they remain limited in two key aspects. 
First, operating within the perspective-view image, each generated view captures only a small portion of the scene due to its restricted field of view.
This often introduces geometric inconsistencies across views, such that generating more views can counterintuitively degrade reconstruction quality  \cite{huang2025gamo}. 
Second, while recent methods attempt to mitigate these inconsistencies through external geometric guidance, depth-conditioned masking \cite{ni2026g4splat}, multi-view conditioning with coordinate embeddings \cite{huang2025gamo}, or positional encodings, they cannot directly steer the diffusion model's internal attention toward structurally relevant observed regions during generation. Consequently, the diffusion process lacks an explicit mechanism to ensure that unobserved regions geometrically align with their assigned structural layout through the model's own feature interactions.

These limitations suggest that panoramic scene completion can provide a natural alternative, as many recent generative world models have increasingly adopted panoramic representations to capture richer global context than narrow perspective images \cite{hyworld22026, omniroam2026, wu360anything}.
Panoramas capture the full spatial layout of a scene in a single representation, making them well suited for modeling global structure \cite{wu2024panodiffusion, sun2019horizonnet}. 
This is especially advantageous for indoor environments, where dominant planar surfaces such as walls, floors, and ceilings define an enclosed geometric envelope \cite{sun2019horizonnet, furukawa2009manhattan, lee2009geometric, liu2019planercnn,zhou2019manhattanwireframes} that can be captured holistically in a single 360$^\circ$ panoramic view.

\paragraph{Main Results:} 

By exposing this omnidirectional global structure to the generative model, panoramic views provide a more coherent context for diffusion-based scene completion.
We present PanoPlane, a geometry-aware novel view synthesis approach that uses panoramic representations for sparse-view indoor reconstruction. Starting from sparse input views, we first train an initial plane-aware 2DGS \cite{Huang2DGS2024} to obtain a coarse geometric reconstruction. 
A vision-language model \cite{Qwen3-VL} detects planar surfaces, such as walls, floors, and ceilings.
From this reconstruction, we render a partially observed equirectangular (ERP) panorama at a selected viewpoint, which contains large missing regions due to sparse input coverage.
We then complete these panoramas using a panoramic flow-matching model \cite{feng2025dit360} that operates in the full latent ERP space to enable global layout awareness during generation. We then guide this completion by our proposed layout anchored attention steering mechanism. 
To determine where attention should be steered, we perform omnidirectional ray tracing from the panoramic camera center and associate each missing pixel with the detected planar surfaces it intersects in the 3D scene. 
This allows completion to follow the room’s underlying layout geometry rather than image space proximity alone, leading to more geometrically consistent scene completion.

Experiments on Replica \cite{straub2019replica}, ScanNet++ \cite{yeshwanth2023scannet++}, and Matterport3D \cite{chang2017matterport3d} show that PanoPlane outperforms prior methods \cite{kerbl20233d, zhu2023FSGS, fan2024instantsplat, xiong2023sparsegs,huang2025gamo,ni2026g4splat}in perceptual  quality across 3, 6, and 9 input views, without any training or per scene fine-tuning of the generative model.
Our contributions are as follows: 

\begin{itemize}

\item We introduce Layout Anchored Attention Steering, a training-free mechanism for sparse view novel view synthesis that guides panoramic scene completion by anchoring the diffusion model's internal attention to the scene's 3D layout. 

\item We propose a semantic-aware plane assignment pipeline that classifies detected planes as layout or non-layout via VLM-based reasoning,  then assigns unobserved regions to layout planes, enabling selective geometric steering.

\item We demonstrate state-of-the-art on three benchmarks with up to +2.71dB PSNR improvement (Replica, 3-view setting), establishing that panoramic priors with layout anchoring provide superior supervision for sparse-view indoor reconstruction.
\end{itemize}

\section{Related Work}
\paragraph{Sparse-View 3DGS}
Early methods regularize 3DGS with monocular depth~\cite{zhu2023FSGS, li2024dngaussian, niemeyer2022regnerf}, SfM-based alignment~\cite{guedon2025matcha}, or epipolar consistency~\cite{zheng2025nexusgs}. Subsequent work extends these constraints to frequency-domain regularization~\cite{zhang2024fregs}, joint depth-normal supervision~\cite{turkulainen2025dnsplatter}, and cross-view geometric alignment~\cite{xiong2023sparsegs, fan2024instantsplat}. Population-level regularization techniques, including stochastic dropout~\cite{park2025dropgaussian}, neural Gaussian priors~\cite{mihajlovic2024splatfields}, and covisibility-aware weighting~\cite{Jang_2025_CVPR}, reduce overfitting but remain fundamentally limited to constraining observed regions. Unlike generative approaches that synthesize novel training views, these geometric regularization methods cannot produce supervision for unobserved scene areas where no input view exists. PanoPlane bridges this gap by combining geometric regularization in observed regions with generative panoramic completion for unobserved areas.

\paragraph{Generative Priors for 3DGS}
To address unobserved regions, recent work injects generative priors into the reconstruction loop. Independent pseudo-view generation~\cite{liu2024deceptive} suffers from cross-view inconsistencies, prompting video-diffusion-based approaches~\cite{zhong2025taming, liu2026reconx} that provide temporally coherent supervision. RI3D~\cite{paliwal2025ri3d} separates repair from inpainting for extreme viewpoints, while G4SPLAT~\cite{ni2026g4splat} lifts monocular depths to metric scale via planar structures. However, methods synthesizing novel camera poses suffer from cumulative drift. OracleGS~\cite{topaloglu2026oraclegs} mitigates this through geometric validation, and GaMO~\cite{huang2025gamo} expands coverage via multi-view outpainting from existing poses. These methods remain confined to fragmented perspective frustums with soft conditioning. PanoPlane shifts to holistic 360° panoramic completion grounded by the scene's 3D planar structure.

\paragraph{Inference-Time Attention Steering}
Recent literature establishes that manipulating a diffusion model's internal attention during inference can improve structural coherence without retraining~\cite{zhang2023controlnet}. Early training-free methods~\cite{hong2023improving, ahn2024self} rely on the assumption that perturbing attention maps isolates structural features, thereby sharpening details and refining sample quality. Subsequent works~\cite{hertz2023prompt} demonstrated that cross-attention governs 2D spatial-to-word relationships, while others~\cite{tumanyan2023plug, cao2023masactrl} exploited self-attention as a driver for transferring 2D pixel compositions across images. Recent advances have sought more granular control: \cite{li2026dcag} demonstrated disentangling Key and Value channels to separate 2D spatial configuration from texture and appearance, \cite{lin2025freecontrol} extracted source attention patterns for 2D structural transfer, and methods like Attentive Eraser~\cite{sun2025attentiveeraser} redirected or masked attention to preserve backgrounds during targeted edits. However these methods lack explicit 3D geometric grounding. To address this limitation, we introduce Layout Anchored attention steering, which upgrades the attention-steering paradigm from 2D image-space heuristics to explicit 3D architectural grounding. Here, we define "layout" strictly as the 3D structural elements of the scene (e.g, walls, floors, and ceilings). Rather than relying on statistical perturbations or 2D spatial masks, our method anchors the self-attention Query-Key interactions directly to analytical plane equations derived from our coarse sparse-view 3DGS reconstruction. We override the model's default 2D proximity bias by deterministically steering unobserved tokens to attend to observed tokens that share the exact same 3D planar room layout.

\section{Our Approach: PanoPlane}
\label{sec:method}

\subsection{Preliminaries}

\paragraph{3D Gaussian Splatting}
3D Gaussian Splatting (3DGS) \cite{kerbl20233d} represents scenes as collections of anisotropic Gaussian primitives optimized via differentiable alpha-blending. We adopt the 2DGS variant \cite{Huang2DGS2024}, which uses oriented planar surfels that provide better surface alignment and enable direct depth and normal supervision. In sparse-view settings, the limited training signal causes collapsed geometry and floater artifacts in unobserved regions.

\paragraph{Flow Matching and Panoramic Diffusion.}
Flow matching~\cite{lipman2023flowmatching} trains a neural network $u_\theta$ to approximate a velocity field that transports samples from a noise distribution to the data distribution along deterministic linear paths. 
Given a clean sample $\mathbf{x}_0$ and Gaussian noise 
$\boldsymbol{\epsilon} \sim \mathcal{N}(\mathbf{0}, \mathbf{I})$, 
intermediate states are defined by the linear interpolation $\mathbf{x}_t = (1-t)\mathbf{x}_0 + t\boldsymbol{\epsilon}$,
and $u_\theta$ is trained to regress the target velocity $\mathbf{v}_t = \boldsymbol{\epsilon} - \mathbf{x}_0$.
At inference time, $u_\theta$ is frozen and clean samples are recovered 
by integrating $u_\theta(\mathbf{x}_t, t)$ from $t{=}1$ (noise) 
to $t{=}0$ (data).

We build on DiT360~\cite{feng2025dit360}, which instantiates $u_\theta$ as a Diffusion Transformer adapted from FLUX~\cite{labs2025flux1kontextflowmatching} for equirectangular 
panoramic generation. The architecture consists of 19 double-stream blocks, in which text and image tokens attend jointly, followed by 38 single-stream blocks that refine 
spatial relationships over the fused token sequence. 
DiT360 applies circular padding at the token level so that the first and last columns of the latent grid are treated as spatially adjacent, allowing self-attention to model the 
$0^\circ/360^\circ$ longitude seam as a continuous boundary. PanoPlane leaves the pretrained weights of DiT360 entirely unchanged, operating solely by modifying the query 
($\mathbf{Q}$) and key ($\mathbf{K}$) matrices within each attention layer at inference time.


\begin{figure}[t]
    \centering
    \includegraphics[width=\linewidth]{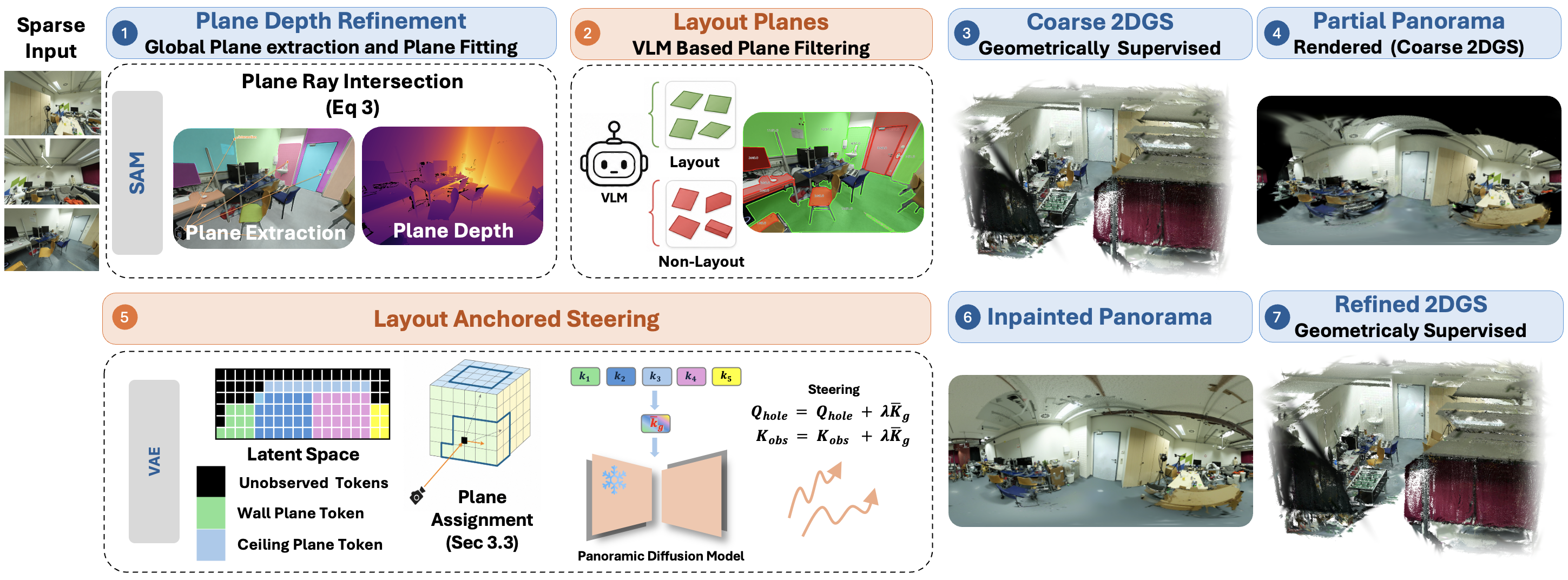}
    \vspace*{-6mm}
    \caption{\textbf{Overview of PanoPlane.} From sparse input views, we recover layout planes, render partial panoramas with holes, assign hole tokens to planes via ray-plane intersection and boundary based assignment, and steer DiT attention to produce geometrically grounded completions. The completed panoramas are converted into cubemap supervision for refined 2DGS reconstruction.}
    \label{fig:overview}
    \vspace*{-4mm}
\end{figure}
\subsection{Plane-Aware Scene Initialization}
\label{sec:method3.2}
\paragraph{Plane Detection and Depth Refinement}

Our plane detection is built upon \cite{ni2026g4splat}. We initialize dense depth maps using Pi3~\cite{wang2026pi3}. We extract per-frame plane instances by intersecting SAM \cite{kirillov2023segany} segmentation masks with surface-normal clusters, merge them into globally consistent 3D planes via point cloud covisibility, and fit plane equations via RANSAC~\cite{fischler1981ransac}. 
For pixels assigned to layout surfaces, noisy per-pixel depth estimates are replaced with depths computed from ray-plane intersection, producing geometrically consistent planar surfaces (Fig.~\ref{fig:overview}, Step~1). Specifically, for a ray originating at camera center $\mathbf{o} \in \mathbb{R}^3$ with direction $\mathbf{d} \in \mathbb{R}^3$, the intersection distance with plane $g$, defined by unit normal $\mathbf{n}_g \in \mathbb{R}^3$ and offset $d_g \in \mathbb{R}$ is given by
\begin{equation}
    L_g = -\frac{\mathbf{n}_g^\top \mathbf{o} + d_g}
               {\mathbf{n}_g^\top \mathbf{d}},
    \label{eq:plane_intersect}
\end{equation}
where $L_g > 0$ indicates a valid intersection in front of the camera. The refined depth for each pixel is then set to $L_g$ of its assigned layout plane.

\paragraph{Semantic Plane Classification}
However, not all detected planes should be treated equally during completion of a unobserved scene. 
We address this by classifying each global plane as \emph{layout} (walls, floors, ceilings) or \emph{non-layout} (furniture, objects) using vision-language-model \cite{Qwen3-VL}. 
For each plane, we render the input image with the plane region lightly highlighted (Fig. \ref{fig:overview}, Step 2), then prompt (see Appendix) the VLM to name the surface. Planes identified as wall, floor, or ceiling are labeled layout; all others are non-layout. Per-frame labels are aggregated by majority vote across all views observing each global plane. During scene completion, only layout planes participate in our attention steering. 

\begin{figure}[t]
    \centering
    \includegraphics[width=0.85\linewidth]{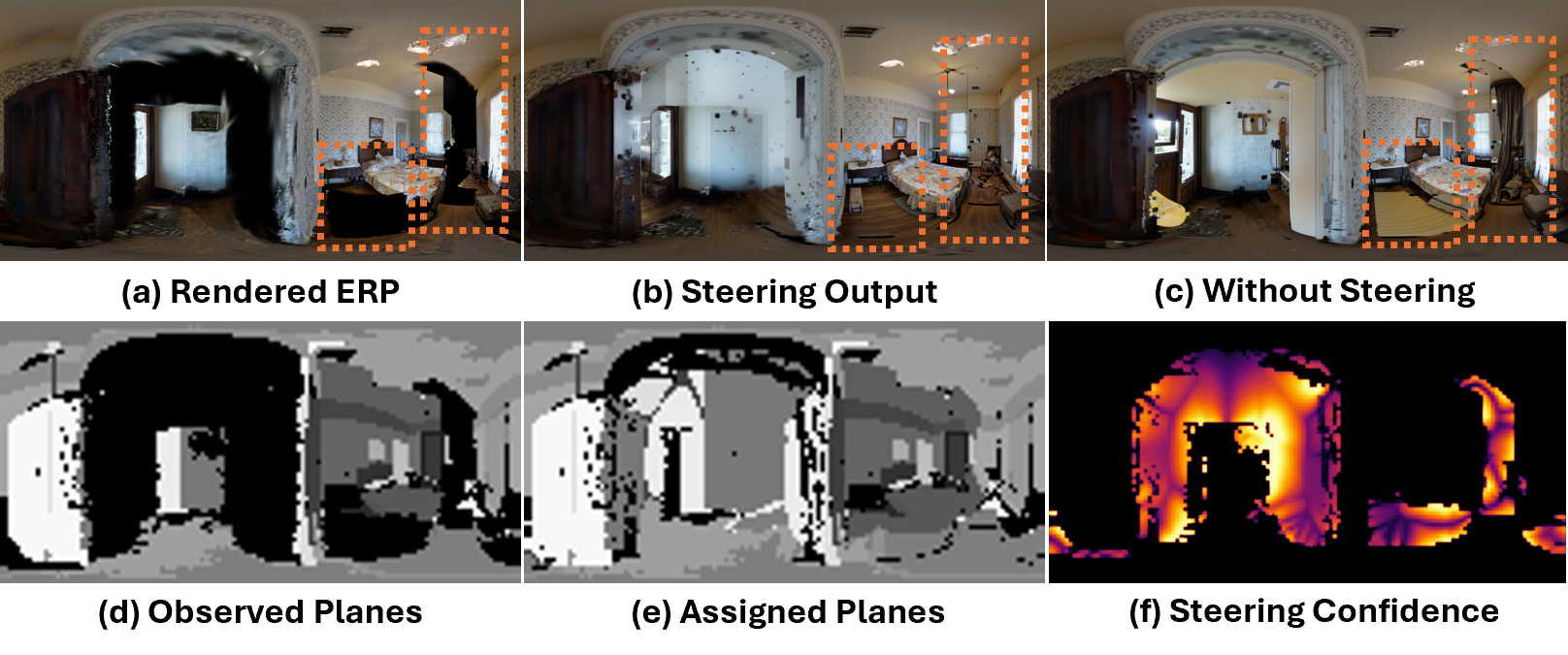}
    \vspace*{-3mm}
    \caption{
        \textbf{Panoramic Completion.}
        (a)~A partially observed equirectangular panorama rendered from the 
        initial 2DGS, with unobserved regions shown in black. 
        (b)~Completion with our layout-anchored steering: unobserved regions 
        are filled as geometrically consistent extensions of the surrounding 
        walls, floor, and ceiling. 
        (c)~Naive panoramic inpainting without steering produces hallucinated 
        surfaces that are geometrically inconsistent with the observed room 
        layout
        (d)~Latent space visualization; each shade of gray 
        denotes observed tokens belonging to a distinct layout plane. 
        (e)~Latent space visualization: each hole token is assigned to its most probable layout plane 
        via the geometric and boundary methods described in Sec.~\ref{sec:completion}, 
        shown here by matching shades. 
        (f)~Assignment confidence map: brighter values indicate higher 
        reliability, with confidence peaking near hole boundaries where 
        spatial context is strongest and decaying toward the interior of 
        large unobserved regions.
        }
        \vspace*{-2mm}
    \label{fig:steering}
\end{figure}
\subsection{Panoramic Completion}
\label{sec:completion}
We complete the scene through 360° panoramic inpainting. This stage proceeds in three steps: we first render ERP panoramas from the coarse 2DGS and select the best candidate for completion; we then determine which layout plane each unobserved region belongs to; finally, we inpaint the panorama using a panoramic flow-matching model with our layout anchored attention steering, guiding the generation of unobserved regions toward the geometric properties of their assigned layout surfaces.

\paragraph{Panorama Rendering and Selection} From the coarse 2DGS, we render six $90^\circ$ cubemap faces at each training camera position and stitch them into equirectangular (ERP) panoramas as shown in Fig.~\ref{fig:steering}a. 
Observed regions are identified by combining the rendered alpha map. The remaining regions form \textbf{holes} that require completion.
The detected global planes and their semantic labels obtained in the previous stage are carried forward into ERP space by projecting (Eq.~\ref{eq:erptopers}) cubemap faces and stitching the resulting per-face planes into a per-pixel plane map in ERP space that identifies which layout surface each observed region belongs to. This map provides the geometric anchors for our downstream hole-plane assignment and attention steering.

To select optimal ERP images for scene completion, we determine scene coverage using a voxel visibility grid following~\cite{ni2026g4splat}, and rank ERP candidates by balancing three criteria: (i)~the number of previously unobserved voxels visible within the hole region, which measures how much new coverage the completion would provide; (ii)~the hole-to-observed ratio $h_i$, which should be large enough to contribute meaningful coverage yet small enough to provide sufficient context for steering; and (iii)~the number of distinct layout planes $P_i$ in the observed region, which determines how many geometric anchors are available. We combine these as
\begin{equation}
    S_i = V_i \cdot f(h_i) \cdot g(P_i),
\end{equation}
where $V_i$ is the number of unobserved voxels visible in the hole region, $f(h_i)$ penalizes panoramas whose hole ratio is either too small, yielding minimal new coverage, or too large, providing insufficient observed context for steering, and $g(P_i)$ rewards layout plane diversity in the observed region. 
Thus, $S_i$ favors panoramas with diverse geometric anchors for stronger scene completion guidance.

\paragraph{Plane Assignment for Hole Tokens.}

To prepare the panorama for inpainting, we encode the rendered ERP image using the VAE into a $64 \times 128$ latent grid, where each token corresponds to a $16 \times 16$ pixel patch. 
The latent grid contains two types of tokens: \emph{observed tokens}, whose corresponding patches contain valid rendered region from the coarse 2DGS, and \emph{hole tokens}, whose patches fall in unobserved regions that require inpainting. 
The ERP-resolution plane map is downsampled to this grid via nearest-neighbor interpolation, so that each observed token inherits the plane ID of its corresponding layout surface, while hole tokens remain unassigned. As shown in Fig.~\ref{fig:steering}d, observed tokens sharing the same plane ID (indicated by the same shade of gray) form contiguous regions corresponding to individual layout surfaces. The challenge is to determine which plane each hole token should be associated with before we inpaint using steering. 
To address this, we employ two complementary strategies and fuse their outputs: geometry based and boundary based assignments.

\paragraph{Geometry Based Hole Assignment}
The geometry based assignment assigns through ray--plane intersection in 3D. For a hole token at latent coordinates $(r,c)$ in a grid of height $H$ and width $W$, we compute a world-space ray direction from the ERP mapping:
\begin{equation}
\mathbf{d}
=
\mathbf{R}
\begin{bmatrix}
\cos\phi \sin\theta \\
-\sin\phi \\
\cos\phi \cos\theta
\end{bmatrix},
\qquad
\theta = \left(\frac{c+0.5}{W}\cdot 2 - 1\right)\pi,
\qquad
\phi = \left(0.5 - \frac{r+0.5}{H}\right)\pi,
\label{eq:erptopers}
\end{equation}
where $\mathbf{R}$ is the camera rotation matrix. 
This ray is intersected with every layout plane and the intersection distance is computed using Eq.~\eqref{eq:plane_intersect}.
The token is assigned to the plane with the smallest positive $L_g$ (Eq.~\eqref{eq:plane_intersect}). The geometric confidence $c_{\text{geo}}$ reflects both the proximity of the hit and the margin over competing planes. Let $L_{\text{best}}$ and $L_{\text{second}}$ denote the intersection distances to the closest and second-closest layout planes, respectively and let $\sigma_L > 0$ be a distance scale parameter that normalizes these quantities into a consistent range:
\begin{equation}
c_{\text{geo}}
=
\exp\!\left(-\frac{L_{\text{best}}}{\sigma_L}\right)
\cdot
\left(
1 - \exp\!\left(-\frac{L_{\text{second}} - L_{\text{best}}}{\sigma_L}\right)
\right).
\end{equation}

\paragraph{Boundary Based Hole Assignment}
The boundary based assignment uses 2D proximity on the latent grid. For each plane, we build a KD-tree over its observed tokens within a few pixels of the hole boundary. Each hole token is assigned to the plane with the nearest boundary tokens in 2D, with confidence defined by the distance and the margin between the closest and second-closest planes:
\begin{equation}
c_{\text{bnd}}
=
\exp\!\left(-\frac{d_{\text{best}}}{\sigma_d}\right)
\cdot
\frac{d_{\text{second}} - d_{\text{best}}}{d_{\text{best}} + \epsilon},
\end{equation}
where $d_{\text{best}}$ and $d_{\text{second}}$ are the 2D euclidean distances to the nearest boundary tokens of the best and second-best planes, respectively and $\sigma_d$ controls the distance falloff. This assignment is strongest near hole boundaries, where spatial context is clear, and weakens for tokens deep inside large holes, where 2D proximity becomes less reliable.

\paragraph{Layout Anchored Attention Steering}
For each hole token, each assignment method contributes a weighted confidence to its selected plane, and the plane with the highest total score is chosen:
\begin{equation}
p_i^\ast
=
\arg\max_g
\left(
w_{\text{geo}} \, c_{\text{geo}}^{(g)}
+
w_{\text{bnd}} \, c_{\text{bnd}}^{(g)}
\right),
\end{equation}
The final confidence (Fig.~\ref{fig:steering}f) for token $i$ is defined as the total weighted score of the winning plane. Non-layout planes are excluded from both assignment methods, so all ray intersections and boundary searches are restricted to the room's layout envelope. We weight both assignment method equally. 

Following \cite{feng2025dit360}, we inpaint the selected panorama and recover the completed image through denoising, while preserving observed region features via token replacement at each step.
Our steering mechanism is integrated into this denoising process.
In the transformer's self-attention, each token's output is a weighted combination of all value vectors, where the weights are determined by the dot products between that token's query $q$ and all other tokens' keys  $k$. By shifting a hole token's query $q$ toward the key $k $ subspace of its assigned plane, we encourage it to attend more strongly to observed tokens from the same surface, thereby promoting structurally consistent completion.

For each layout plane $g$ with observed token set $O_g$ and hole token set $H_g$, we compute a plane centroid by averaging the key vectors of the observed tokens. We then apply two complementary modifications: each hole token's query is shifted toward the plane centroid, and the observed tokens on that plane have their keys reinforced, making them stronger attention targets:
%
\begin{equation}
\begin{aligned}
\bar{\mathbf{k}}_g
&= \frac{1}{|O_g|} \sum_{j \in O_g} \mathbf{k}_j,
\qquad
\begin{aligned}
\mathbf{q}_i &\leftarrow \mathbf{q}_i + \lambda \,\bar{\mathbf{k}}_g,
&\quad i \in H_g \\
\mathbf{k}_j &\leftarrow \mathbf{k}_j + \lambda \,\bar{\mathbf{k}}_g,
&\quad j \in O_g
\end{aligned}
\end{aligned}
\label{eq:plane_steering}
\end{equation}

Steering is applied to single-stream blocks 10–37 of DiT360~\cite{feng2025dit360}, which empirical analysis shows encode geometric surface structure (Appendix~\ref{app:A}); double-stream and early blocks are unchanged.
Steering is active only when the denoising timestep $t$ exceeds a threshold $\tau$, corresponding to the early phase in which the global spatial layout is established, and is disabled in later steps, when the model refines texture and fine detail. All modifications are applied directly to the $\mathbf{Q}$ and $\mathbf{K}$ matrices in the attention layer and remain fully compatible with standard FlashAttention \cite{dao2022flashattention}, requiring neither custom attention masks nor additional memory overhead.

\subsection{Refinement}
Each completed ERP image (Fig.\ref{fig:steering}b) is decomposed into six $90^\circ$ cubemap faces standard ERP-to-cubemap reprojection and added to the training set. The plane-aware pipeline
(Sec.\ref{sec:method3.2}) is re-executed over this expanded view set, and the 2DGS model is
refined using both sources of supervision. For original input views, we apply the standard photometric loss:
\begin{equation}
\mathcal{L}_{\text{input}}
=
(1-\lambda_s)\,\mathcal{L}_1(\hat{I}_i, I_i)
+
\lambda_s\,\mathcal{L}_{\text{D-SSIM}}(\hat{I}_i, I_i).
\end{equation}

For completed views, the same objective is downweighted by
$\alpha_c = 0.01$ to prevent synthetic supervision from overpowering real
observations:
\begin{equation}
\mathcal{L}_{\text{complete}}
=
\alpha_c
\left[
(1-\lambda_s)\,\mathcal{L}_1(\hat{C}_j, C_j)
+
\lambda_s\,\mathcal{L}_{\text{D-SSIM}}(\hat{C}_j, C_j)
\right].
\end{equation}

All views share geometric regularization against the plane-refined depth maps:
\begin{equation}
\mathcal{L}_{\text{geo}}
=
\lambda_d\,\mathcal{L}_{\text{depth}}
+
\lambda_n\,\mathcal{L}_{\text{normal}}
+
\lambda_\kappa\,\mathcal{L}_{\text{curv}},
\end{equation}
applied at full strength regardless of view source to keep Gaussian positions
anchored to the planar geometry.
\section{Results and Experiments}
\label{sec:experiments}
\subsection{Experiments Settings}
\paragraph{Datasets, Metrics}
We evaluate on three standard indoor benchmarks spanning different scene complexity, totaling 15 scenes across 21 experimental configurations. 
Matterport3D \cite{chang2017matterport3d} provides building-scale indoor environments captured as panoramic RGB-D views at multiple positions throughout each scene. We extract six 90° cubemap faces from each panoramic viewpoint, producing a pool of perspective images with known camera poses per room. 
From this pool, we randomly select perspective views as sparse training inputs, ensuring they originate from different panoramic positions. 
We evaluate on 6 room-level scenes with 6 input views. Held-out perspective views from the same room serve as the test set for evaluation. 
Replica \cite{straub2019replica} provides photorealistic synthetic indoor scenes with ground truth geometry. We evaluate on 3 scenes across 3, 6, and 9 input views. ScanNet++ \cite{yeshwanth2023scannet++} provides real-world indoor scans captured with a DSLR camera, featuring challenging lighting and cluttered layouts. From each scene's image sequence, we randomly select 6 images as sparse training views and a separate set of 100 images as held-out test views. We report three standard novel view synthesis metrics: PSNR, SSIM \cite{wang2004ssim}, and LPIPS \cite{zhang2018lpips}. All metrics are computed on held-out test views that were not used during reconstruction.

\paragraph{Baselines}

We compare PanoPlane against six baselines from three categories: 3DGS~\cite{kerbl20233d}; regularization-based methods, including SparseGS~\cite{xiong2023sparsegs}, FSGS~\cite{zhu2023FSGS}, and InstantSplat~\cite{fan2024instantsplat}; and generative-prior methods, including GaMO~\cite{huang2025gamo} and G4Splat~\cite{ni2026g4splat}. All methods use their default settings from the original papers.



\subsection{Main Improvements}
\begin{figure}[t]
    \centering
    \includegraphics[width=\linewidth]{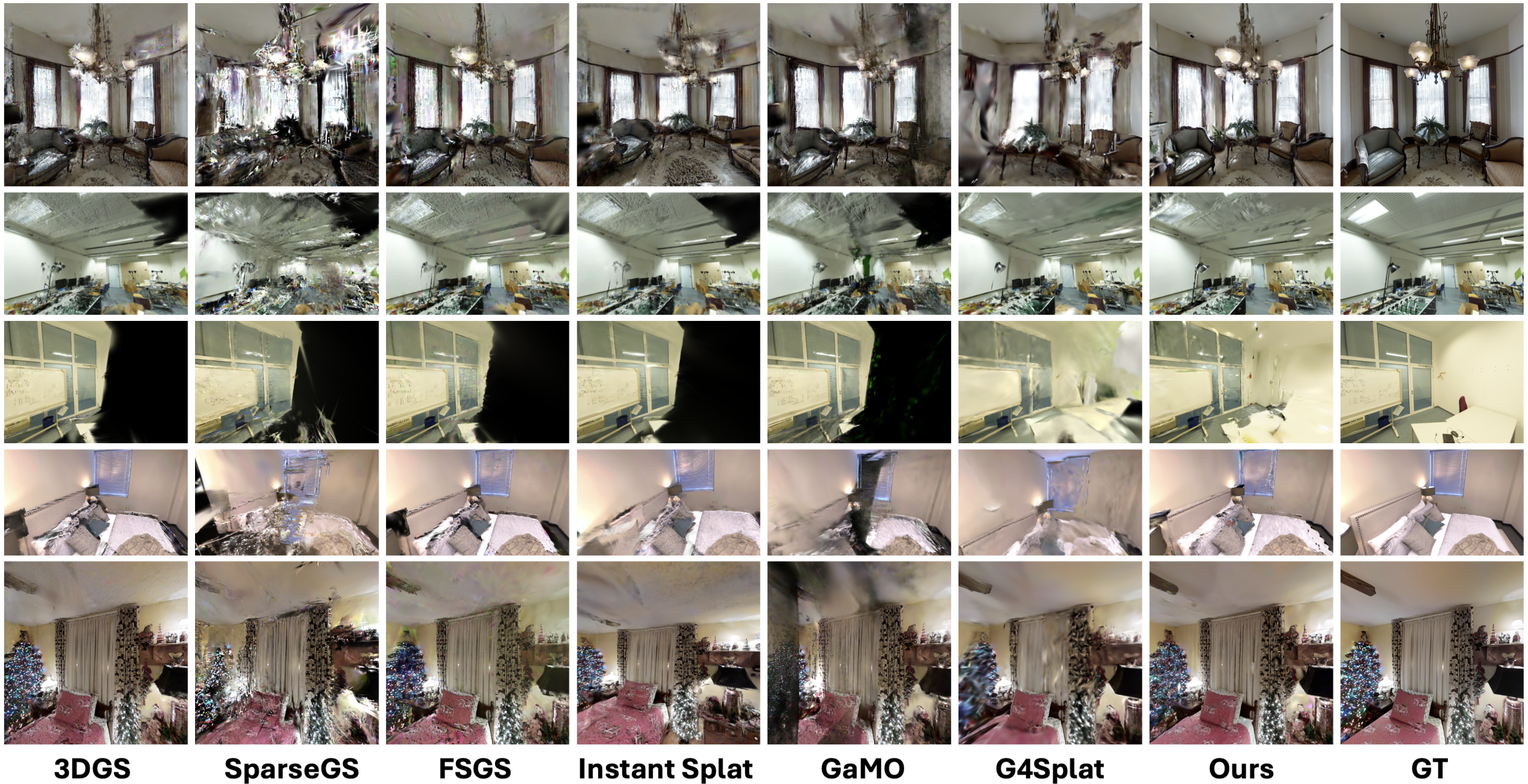}
    \vspace*{-5mm}
    \caption{\textbf{Qualitative comparison on Matterport3D, Replica and ScanNet++} Each row shows a challenging scene with large unobserved regions. Perspective-based methods (3DGS, SparseGS, FSGS, InstantSplat) artifacts in wide baseline views. Generative-prior methods partially recover scenes but still struggle with color bleeding and over-smooth textures. PanoPlane (ours) recovers geometrically consistent surfaces across all unobserved regions, correctly extending walls, floors, and ceilings that other methods fail to reconstruct.  }
    \label{fig:qualitative_results}
\end{figure}

Tables~\ref{tab:main_dataset_comparison} and Figure~\ref{fig:qualitative_results} present quantitative and qualitative comparisons across all three benchmarks. 
PanoPlane achieves the highest state-of-the-art performance across all datasets and view counts. On primary 6-view setting, PanoPlane outperforms strongest baseline (\cite{ni2026g4splat}) by +2.60dB PSNR on Matterport3D, +1.75dB on Replica and +1.90dB on ScanNet++. These gains are consistent across all three perceptual metrics.
Fig.\ref{fig:qualitative_results} illustrates failure modes of existing methods. G4Splat completes unobserved regions but its perspective inpainting introduces color bleeding and texture smoothing into adjacent observed regions, degrading reconstruction quality even where ground-truth supervision is available. GaMO fails to produce coherent reconstructions when input views have wide baselines, and in some cases corrupts already observed regions. In scenes with strong occlusions (Replica), our method resolves hidden surfaces that perspective methods cannot recover.
Table~\ref{tab:main_dataset_comparison} also shows results on Replica under 3 and 9 input views. PanoPlane's advantage is largest under the most extreme sparsity (3 views, +2.71 dB over best baseline), where the global context provided by panoramic completion is most valuable. 
At 9 views, PanoPlane still achieves the best performance (+0.25 dB over G4Splat), indicating that layout steering provides geometric consistency benefits even when input coverage is relatively dense.
\begin{table*}[t]
\centering
\scriptsize
\caption{Average comparison on Matterport3D~\cite{chang2017matterport3d}, Replica~\cite{straub2019replica}, and ScanNet++~\cite{yeshwanth2023scannet++}. Best, second best, and third best results are highlighted in red, orange, and yellow, respectively.}
\vspace*{-2mm}
\label{tab:main_dataset_comparison}
\resizebox{1.0\textwidth}{!}{
\begin{tabular}{l ccc ccc ccc|| ccc ccc}
\toprule
\multirow{2}{*}{\textbf{Method}} 
& \multicolumn{3}{c}{\textbf{Matterport3D (6 views)}} 
& \multicolumn{3}{c}{\textbf{ScanNet++ (6 views)}} 
& \multicolumn{3}{c||}{\textbf{Replica (6 views)}} 
& \multicolumn{3}{c}{\textbf{Replica (3 views)}}
& \multicolumn{3}{c}{\textbf{Replica (9 views)}}\\
\cmidrule(lr){2-4}
\cmidrule(lr){5-7}
\cmidrule(lr){8-10}
\cmidrule(lr){11-13}
\cmidrule(lr){14-16}
& PSNR$\uparrow$ & SSIM$\uparrow$ & LPIPS$\downarrow$ 
& PSNR$\uparrow$ & SSIM$\uparrow$ & LPIPS$\downarrow$ 
& PSNR$\uparrow$ & SSIM$\uparrow$ & LPIPS$\downarrow$ 
& PSNR$\uparrow$ & SSIM$\uparrow$ & LPIPS$\downarrow$
& PSNR$\uparrow$ & SSIM$\uparrow$ & LPIPS$\downarrow$ \\
\midrule

3DGS~\cite{kerbl20233d}
& 13.83 & 0.46 & \cellcolor{yellow!35}0.50 
& \cellcolor{yellow!35}13.14 & 0.55 & \cellcolor{orange!35}0.45
& 16.25 & 0.67 & \cellcolor{yellow!35}0.30 
& 10.26 & \cellcolor{yellow!35}0.43 & \cellcolor{red!35}0.36
& 18.77 & 0.71 & \cellcolor{red!35}0.26 \\

FSGS~\cite{zhu2023FSGS}
& 13.59 & \cellcolor{yellow!35}0.49 & \cellcolor{yellow!35}0.51 
& 13.61 & 0.60 & 0.48
& 16.61 & 0.71 & 0.31 
& \cellcolor{orange!35}10.78 & \cellcolor{orange!35}0.49 & \cellcolor{orange!35}0.38
& \cellcolor{yellow!35}19.08 & \cellcolor{yellow!35}0.74 & 0.30 \\

SparseGS~\cite{xiong2023sparsegs}
& 12.69 & 0.36 & 0.55 
& \cellcolor{yellow!35}13.80 & 0.52 & 0.48
& 13.81 & 0.53 & 0.45 
& \cellcolor{yellow!35}10.57 & 0.38 & 0.49
& 16.24 & 0.62 & 0.40 \\

InstantSplat~\cite{fan2024instantsplat}
& 12.63 & 0.45 & 0.53 
& 12.68 & \cellcolor{yellow!35}0.59 & \cellcolor{yellow!35}0.46
& 14.68 & 0.64 & 0.38 
& 7.62 & 0.23 & 0.52
& 16.13 & 0.69 & 0.31 \\

GaMO~\cite{huang2025gamo}
& 12.57 & 0.41 & 0.59 
& 11.01 & 0.44 & 0.58
& 12.05 & 0.50 & 0.56 
& 8.28 & 0.27 & 0.66
& 14.69 & 0.60 & 0.44 \\

G4Splat~\cite{ni2026g4splat}
& \cellcolor{orange!35}14.59 & \cellcolor{orange!35}0.52 & \cellcolor{yellow!35}0.51 
& \cellcolor{yellow!35}13.86 & 0.53 & 0.51
& 17.80 & \cellcolor{yellow!35}0.74 & 0.37 
& -- & -- & --
& \cellcolor{orange!35}21.81 & \cellcolor{orange!35}0.77 & \cellcolor{yellow!35}0.29 \\


\textbf{PanoPlane}
& \cellcolor{red!35}17.19 & \cellcolor{red!35}0.53 & \cellcolor{red!35}0.44 
& \cellcolor{red!35}15.76 & \cellcolor{red!35}0.66 & \cellcolor{red!35}0.42 
& \cellcolor{red!35}19.55 & \cellcolor{red!35}0.78 & \cellcolor{red!35}0.29 
& \cellcolor{red!35}13.49 & \cellcolor{red!35}0.58 & \cellcolor{yellow!35}0.44
& \cellcolor{red!35}22.06 & \cellcolor{red!35}0.80 & \cellcolor{orange!35}0.27 \\

\bottomrule
\end{tabular}
}
\end{table*}

\subsection{Ablation Studies}

\paragraph{Layout Steering}
As observed in Table~\ref{tab:ablation_perdataset}, layout steering is the single largest contributor. Removing it entirely and replacing it with naive DiT360 inpainting causes the largest single drop ($-1.02$ dB average PSNR), confirming that the Q/K modification described in Eq.~\ref{eq:plane_steering} is the primary mechanism driving geometric consistency in the completed panoramas.

\paragraph{Semantic plane classification} We observe that semantic plane classification is necessary for steering to work. Row (e) in Table~\ref{tab:ablation_perdataset} shows that applying boundary-based assignment without semantic filtering achieves 16.05 dB, identical to disabling steering entirely (16.07 dB). This is because naive assignment directs unobserved tokens toward non-layout surfaces (furniture, objects) as often as toward walls and ceilings, causing the steering to hallucinate flat textures rather than enforce geometric consistency. Semantic filtering resolves this by restricting the anchor set to structurally meaningful surfaces.

\newcommand{\cmark}{\ding{51}}
\newcommand{\xmark}{\ding{55}}

\begin{table*}[t]
\centering
\caption{\textbf{Ablation} Results when each pipeline component is individually removed, averaged over all 15 scenes. 
LS: Layout Steering, BND: Boundary assignment, GEO: Geometric assignment, SF: Semantic Filtering. 
Best, second best, and third best results are highlighted in red, orange, and yellow, respectively.
Averages computed over the same 15 scenes.}
\label{tab:ablation_perdataset}
\resizebox{0.95\textwidth}{!}{
\begin{tabular}{ccccc ccc ccc ccc ccc}
\toprule
& & & & & \multicolumn{3}{c}{\textbf{Matterport3D}} 
& \multicolumn{3}{c}{\textbf{Replica (6-view)}} 
& \multicolumn{3}{c}{\textbf{ScanNet++}} 
& \multicolumn{3}{c}{\textbf{Average}} \\
\cmidrule(lr){6-8} \cmidrule(lr){9-11} \cmidrule(lr){12-14} \cmidrule(lr){15-17}
& LS & BND & GEO & SF 
& PSNR$\uparrow$ & SSIM$\uparrow$ & LPIPS$\downarrow$ 
& PSNR$\uparrow$ & SSIM$\uparrow$ & LPIPS$\downarrow$ 
& PSNR$\uparrow$ & SSIM$\uparrow$ & LPIPS$\downarrow$ 
& PSNR$\uparrow$ & SSIM$\uparrow$ & LPIPS$\downarrow$ \\
\midrule
(f) & \xmark & -- & -- & -- 
& 16.91 & 0.53 & 0.44 
& 17.39 & 0.75 & 0.31 
& 14.57 & 0.63 & 0.44 
& 16.07 & 0.613 & 0.415 \\
(e) & \cmark & \cmark & \xmark & \xmark 
& 16.14 & \cellcolor{yellow!35}0.53 & 0.45 
& \cellcolor{yellow!35}19.51 & \cellcolor{yellow!35}0.78 & \cellcolor{yellow!35}0.29 
& 14.24 & 0.62 & 0.45 
& 16.05 & 0.616 & 0.417 \\
(d) & \cmark & \xmark & \cmark & \cmark 
& 16.93 & 0.53 & 0.44 
& 18.69 & 0.77 & 0.30 
& \cellcolor{yellow!35}15.23 & 0.64 & 0.44 
& 16.60 & \cellcolor{yellow!35}0.623 & \cellcolor{yellow!35}0.409 \\
(c) & \cmark & \cmark & \xmark & \cmark 
& \cellcolor{yellow!35}16.97 & 0.53 & \cellcolor{yellow!35}0.44 
& \cellcolor{red!35}19.90 & \cellcolor{red!35}0.78 & \cellcolor{red!35}0.29 
& \cellcolor{orange!35}15.39 & \cellcolor{orange!35}0.65 & \cellcolor{orange!35}0.43 
& \cellcolor{orange!35}16.92 & \cellcolor{orange!35}0.629 & \cellcolor{orange!35}0.403 \\
(b) & \cmark & \cmark & \cmark & \xmark 
& \cellcolor{orange!35}17.10 & \cellcolor{orange!35}0.53 & \cellcolor{red!35}0.44 
& 19.24 & 0.71 & 0.33 
& 15.13 & \cellcolor{yellow!35}0.65 & \cellcolor{yellow!35}0.43 
& \cellcolor{yellow!35}16.74 & 0.613 & 0.413 \\
(a) & \cmark & \cmark & \cmark & \cmark 
& \cellcolor{red!35}\textbf{17.19} & \cellcolor{red!35}\textbf{0.53} & \cellcolor{orange!35}\textbf{0.44} 
& \cellcolor{orange!35}\textbf{19.55} & \cellcolor{orange!35}\textbf{0.78} & \cellcolor{orange!35}\textbf{0.29} 
& \cellcolor{red!35}\textbf{15.76} & \cellcolor{red!35}\textbf{0.66} & \cellcolor{red!35}\textbf{0.42} 
& \cellcolor{red!35}\textbf{17.09} & \cellcolor{red!35}\textbf{0.632} & \cellcolor{red!35}\textbf{0.400} \\
\bottomrule
\end{tabular}
}
\end{table*}



\begin{table}[t]
\centering
\setlength{\tabcolsep}{3pt}
\renewcommand{\arraystretch}{0.92}

\begin{minipage}{0.28\linewidth}
\centering

\scriptsize
\vspace{-6mm}
\caption{Q and K Steering}
\label{tab:qk_steering}
\begin{tabular}{lccc}
\toprule
Config. & PSNR$\uparrow$ & SSIM$\uparrow$ & LPIPS$\downarrow$ \\
\midrule
Both Q,K & 17.09 & 0.63 & 0.40 \\
Q only   & 16.69 & 0.65 & 0.39 \\
K only   & 15.63 & 0.61 & 0.42 \\
\bottomrule
\end{tabular}

\vspace{-2mm}

\scriptsize
\caption{Run Times (mins)}
\label{tab:runtime}
\begin{tabular}{lc}
\toprule
Method & Time$\downarrow$ \\
\midrule
3DGS~\cite{kerbl20233d}      & 17  \\
FSGS~\cite{zhu2023FSGS}      & 24  \\
SparseGS~\cite{xiong2023sparsegs}  & 23  \\
GaMO~\cite{huang2025gamo}      & 38  \\
G4Splat~\cite{ni2026g4splat}   & 120 \\
\textbf{Panoplane} & 39  \\
\bottomrule
\end{tabular}

\end{minipage}
\begin{minipage}{0.66\linewidth}
\centering

\hspace*{-12mm}%
\begin{minipage}{0.9\linewidth}
\raggedright
\begin{minipage}{0.60\linewidth}
\centering
\includegraphics[width=0.95\linewidth]{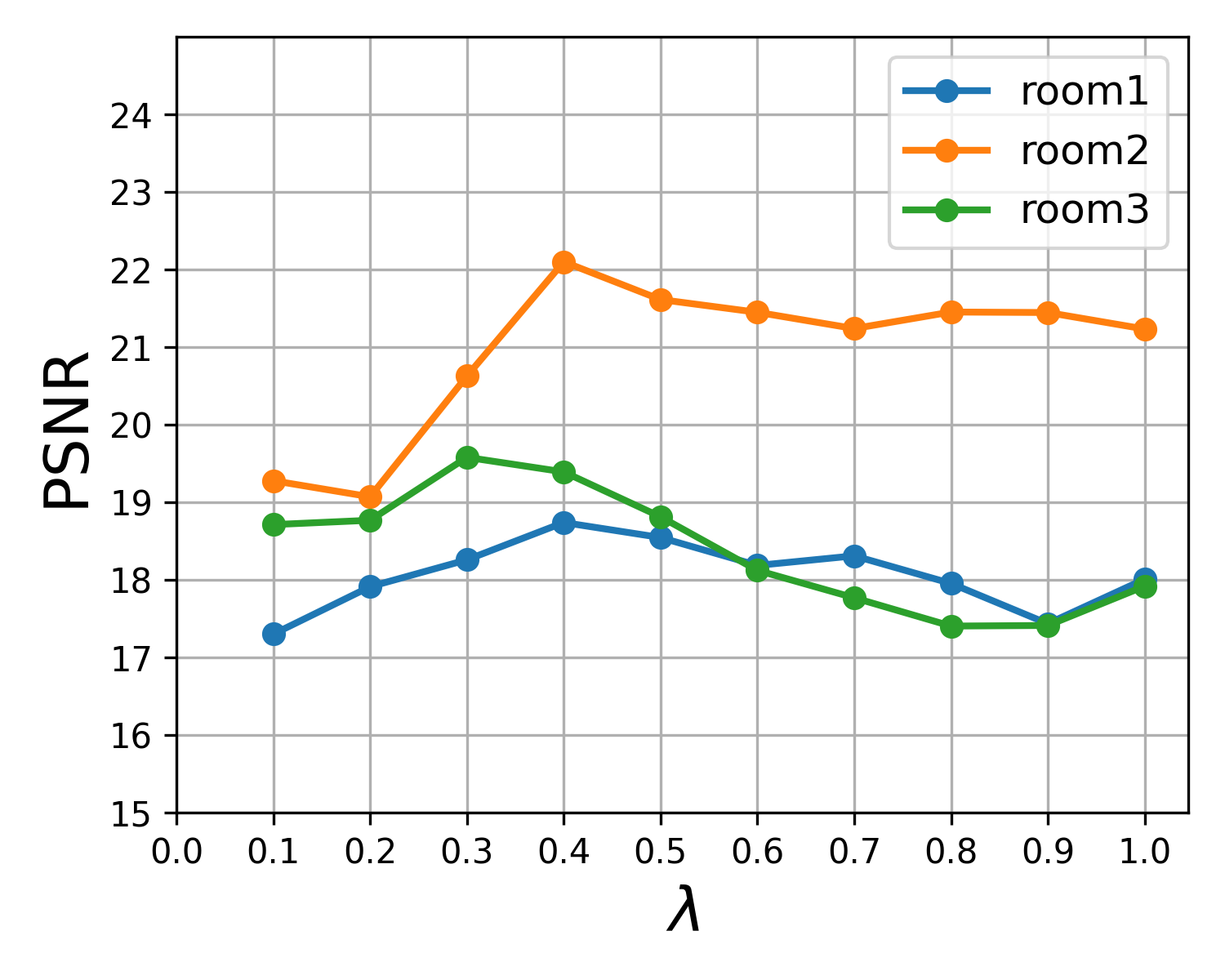}
\end{minipage}%
\hspace{1.5mm}
\begin{minipage}{0.36\linewidth}
\scriptsize
\makebox[0pt][l]{%
  \hspace*{-2mm}%
  \begin{minipage}{1.4\linewidth}
  \raggedright
  \captionof{figure}{
  \textbf{Analysis} PSNR vs steering strength $\lambda$ on Replica. Performance peaks at $\lambda=0.4$; stronger steering over-constrains the model, producing flat textureless completions.} 

  \label{fig:psnr}
  \end{minipage}}

\end{minipage}
\end{minipage}

\end{minipage}

\vspace*{-8mm}
\end{table}
\paragraph{Geometric and Boundary-Based Assignments} Boundary assignment alone (row c, 16.92 dB) outperforms geometric assignment alone (row d, 16.60 dB), though the two together (row a, 17.09 dB) provide the best result. The benefit of combining both is most pronounced on scenes with large holes spanning multiple layout planes: boundary assignment is reliable near hole edges where spatial context is strong, but degrades toward hole interiors where 2D proximity becomes ambiguous. Geometric ray-plane intersection provides stable assignments in those interior regions, explaining why neither method alone matches their combination.

\paragraph{Steering Strength} Figure~\ref{fig:psnr} plots PSNR as a function of $\lambda$ in Eq.~\ref{eq:plane_steering} across Replica dataset. Performance peaks at $\lambda= 0.4$ for Replica scenes. Values above $\lambda > 0.4$ causes over-steering of the hole tokens which causes the inpainting to collapse and producing flat, texture less completions. We use $\lambda = 0.4$ for all reported results.

\paragraph{Q and K Steering} Table \ref{tab:qk_steering} shows that modifying both Q and K outperforms modifying Q alone or K alone across all three datasets. Shifting Q toward the plane centroid encourages the hole token to attend towards its assigned surface. K only steering without Q adjustment yields the largest drop, suggesting that query alignment is the dominant mechanism.

\paragraph{Discussion}
Table~\ref{tab:ablation_perdataset} shows that semantic classification and geometric assignment each improve performance (+0.87dB and +0.53dB respectively), with overlapping benefits yielding a combined gain of +1.02dB. Most revealingly, boundary assignment without semantic filtering (16.05dB) performs on par with disabling steering entirely (16.07dB): naive assignment directs tokens toward non-layout surfaces as often as correct ones, negating any benefit. Geometric assignment is more robust to this failure (16.74dB without filtering), while semantic classification further improves boundary assignment (16.92dB) by excluding non-layout planes. The best result (17.09dB) requires both: semantic classification determines which planes to steer toward, geometric assignment determines where, and steering adds negligible runtime overhead (Table~\ref{tab:runtime}).

\section{Conclusion, Limitations, and Future Work}
\label{conclusion}
We presented Panoplane, a training-free approach to sparse-view indoor 3DGS reconstruction that anchors panoramic diffusion inpainting to the scene's 3D planar structure, steering the transformer's internal attention so that unobserved regions are completed as geometric extensions of detected layout surfaces rather than unconstrained hallucinations. Experiments on three standard benchmarks demonstrate state-of-the-art reconstruction quality without any fine-tuning of the generative model. Our method assumes detectable planar structure and may underperform in scenes dominated by curved surfaces or heavy clutter that occludes the structural envelope; the VLM classification can also misclassify ambiguous surfaces such as built-in shelving or large mirrors. Future directions include adaptive per-head steering strengths matched to each attention head's geometric sensitivity, direct injection of plane-derived depth into the 3DGS training loss for completed views, and extension to non-planar geometric primitives to handle a broader range of indoor environments.
\bibliographystyle{plainnat}
\bibliography{references}

\clearpage
\appendix
\section*{Appendix}
\addcontentsline{toc}{section}{Appendix}
In this appendix, we provide additional discussion, experimental results, and technical details: implementation details (Sec.~\ref{app:A}), Failure cases (Sec.~\ref{app:failures}), and additional qualitative and quantitative results (Sec.~\ref{app:qual}, ~\ref{app:quant}).

\section{Implementation Details}
\label{app:A}

\paragraph{General Configuration}
All experiments are conducted on a NVIDIA A6000 single GPU. The initial 2DGS~\cite{Huang2DGS2024} is trained for 7,000 iterations before the panoramic completion, with refinement of 7000 iterations as well.

\paragraph{VLM Prompting for Plane Classification.}
We use Qwen3-VL-3B~\cite{Qwen3-VL} for semantic plane classification 
with chain-of-thought prompting. For each detected plane, we render 
the input image with a red contour outlining the plane boundary and 
a numeric identifier at the region's centroid. The VLM receives the 
following prompt:

\begin{quote}
\small
\texttt{Look at the region outlined in red and marked `\{id\}' in 
this indoor room photo.\\
Think step by step:\\
1. What does this region look like?\\
2. Is this region located on a wall, floor, ceiling or some other 
surface?\\
3. Give your final answer as a single word: wall, floor, ceiling, 
bed, table, shelf, cabinet, window, door, or other.}
\end{quote}

The model generates up to 200 tokens of reasoning. We parse the 
response by finding the \emph{last} occurrence of any label keyword; 
if the final keyword is ``wall'', ``floor'', or ``ceiling'', the 
plane is labeled \emph{layout}, otherwise \emph{non-layout}. Using 
the last occurrence rather than the first ensures the final 
conclusion of the chain-of-thought reasoning takes precedence over 
intermediate mentions. Planes whose bounding box is smaller than 
$32 \times 32$ pixels in either dimension are automatically labeled 
non-layout without querying the VLM. Classification adds 
approximately 10 seconds per input frame.

\paragraph{Layer Selection for Steering}
To determine which transformer layers should receive steering, we 
compute a per-layer plane affinity ratio. For layer $\ell$, we sample 
a set $S$ of hole tokens and measure how strongly each hole token 
attends to observed tokens on the same plane versus a different plane:
\begin{equation}
r_\ell = \frac{1}{|S|} \sum_{i \in S} 
\frac{\mathbf{q}_i^\top \bar{\mathbf{k}}_{\text{same}}}
{\mathbf{q}_i^\top \bar{\mathbf{k}}_{\text{other}} + \epsilon},
\end{equation}
where $\mathbf{q}_i$ is the query vector of hole token $i$ at layer 
$\ell$, $\bar{\mathbf{k}}_{\text{same}}$ is the mean key vector of 
observed tokens belonging to the same plane as token $i$, 
$\bar{\mathbf{k}}_{\text{other}}$ is the mean key vector of observed 
tokens on a randomly selected different plane, and $\epsilon$ is a 
small constant for numerical stability. We evaluate $r_\ell$ at 
multiple denoising timesteps ($t \in \{0.2, 0.4, 0.6, 0.8\}$). 
Layers 0--9 exhibit unstable $r_\ell$ that fluctuates across 
timesteps, indicating they handle global properties rather than 
geometric surface grouping. Layers 10--37 consistently produce 
$r_\ell > 1.5$, confirming they naturally cluster tokens by layout 
surface. We restrict steering to these layers, amplifying an existing 
geometric signal rather than imposing a foreign one.

\paragraph{Panoramic Inpainting}
We run DiT360~\cite{feng2025dit360} with 50 denoising steps per panorama. The observed content is encoded via RF-inversion~\cite{rout2025semantic} with token replacement 
strength $\tau_{\text{replace}} = 0.90$ and text guidance scale 2.8. 
A generic text prompt (``This is a panoramic image of an indoor 
room.'') is used for all scenes. Inpainting a single 
$512 \times 1024$ ERP panorama takes approximately 2 minutes.
\paragraph{Baselines Configurations}
All baselines are evaluated using their official code repositories 
and default configurations. 3DGS~\cite{kerbl20233d} is trained 
directly on sparse inputs with default densification and pruning. 
FSGS~\cite{zhu2023FSGS} uses its default depth-guided Gaussian 
unpooling. SparseGS~\cite{xiong2023sparsegs} uses its default depth 
and appearance regularization. 
InstantSplat~\cite{fan2024instantsplat} uses 
DUSt3R~\cite{wang2024dust3r} for stereo initialization with default 
bundle adjustment. GaMO~\cite{huang2025gamo} uses its default 
multi-view outpainting with geometry-aware denoising. 
G4SPLAT~\cite{ni2026g4splat} uses 
MASt3R-SfM~\cite{duisterhof2025mastrsfm} for chart-aligned depth 
estimation and See3D~\cite{Ma2025See3D} for video diffusion 
inpainting. No hyperparameter tuning was performed for any baseline. 
All methods receive the same input images and test views per scene.

\section{Failure Cases and Limitations}
\label{app:failures}
\vspace{-7mm}
\begin{figure}[H]
\centering
\includegraphics[width=\textwidth]{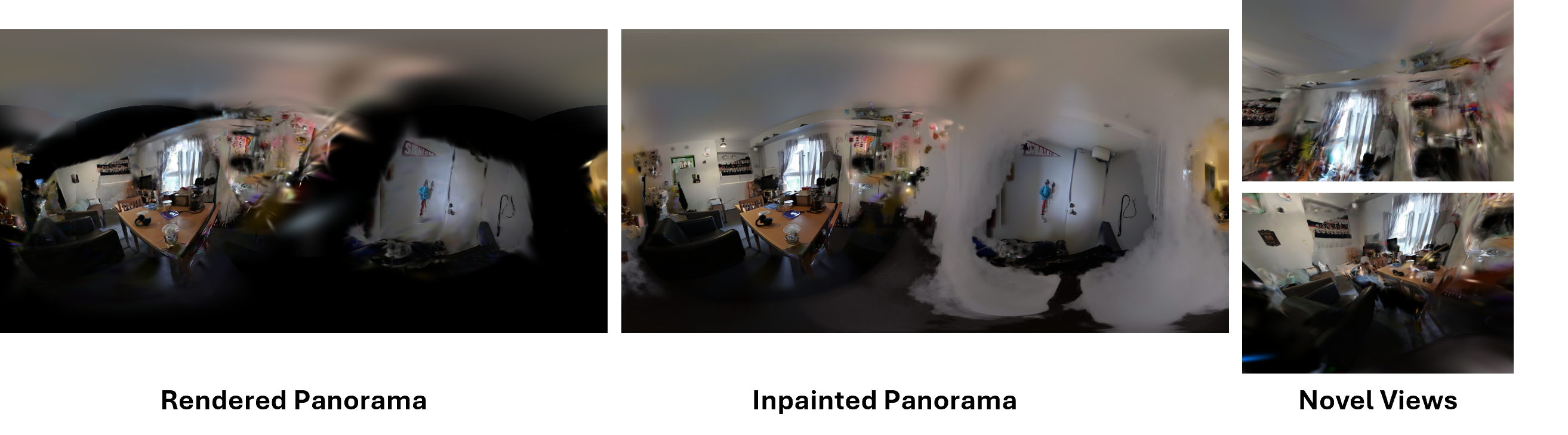}
\caption{\textbf{Failure case:} When the rendered 2DGS panorama contains significant artifacts, DiT360 fails to produce coherent inpainting and layout-anchored steering worsens the result by reinforcing incorrect plane assignments.}
\label{fig:failure}
\end{figure}

While PanoPlane requires no per-scene fine-tuning or training of 
the diffusion model, it inherits the limitations of the underlying 
foundation model. DiT360~\cite{feng2025dit360} is trained on 
large-scale panoramic datasets, and its generative quality degrades 
when the input panorama deviates from its training distribution, 
producing blurry or semantically incoherent completions regardless 
of the steering applied. First, rooms with unusual geometry or non-residential environments 
underrepresented in the training data. Second, when the coarse 2DGS reconstruction contains significant artifacts, incorrect depth, or misaligned 
planes, the rendered ERP panorama no longer resembles a natural 
indoor scene. This distribution shift can trigger attention collapse 
during denoising, where the model's attention concentrates on a 
small subset of tokens, producing repetitive textures or flat 
featureless regions (Fig.~\ref{fig:failure}). Incorrect plane 
assignments push hole tokens toward the wrong surface, and the 
joint Q/K modification amplifies the error.

\section{Additional Quantitative Results}
\label{app:quant}

\begin{table*}[!ht]
\centering
\caption{\textbf{Per-scene results on Matterport3D (6 views).} 
Best, second best, and third best are highlighted in 
\colorbox{red!35}{red}, \colorbox{orange!35}{orange}, and \colorbox{yellow!35}{yellow}.}
\label{tab:mp3d_perscene}
\setlength{\tabcolsep}{4pt}

\vspace{1mm}
\resizebox{\textwidth}{!}{
\begin{tabular}{l ccc ccc ccc}
\toprule
\multirow{2}{*}{\textbf{Method}} 
& \multicolumn{3}{c}{\textbf{2t7WUuJeKo (2)}} 
& \multicolumn{3}{c}{\textbf{RPmz2sHmrrY (2)}} 
& \multicolumn{3}{c}{\textbf{YVUC4YcDtcY}} \\
\cmidrule(lr){2-4} \cmidrule(lr){5-7} \cmidrule(lr){8-10}
& PSNR$\uparrow$ & SSIM$\uparrow$ & LPIPS$\downarrow$ 
& PSNR$\uparrow$ & SSIM$\uparrow$ & LPIPS$\downarrow$ 
& PSNR$\uparrow$ & SSIM$\uparrow$ & LPIPS$\downarrow$ \\
\midrule
3DGS~\cite{kerbl20233d}
& \cellcolor{orange!35}13.18 & 0.46 & \cellcolor{orange!35}0.47 
& 12.89 & 0.49 & \cellcolor{yellow!35}0.49 
& \cellcolor{yellow!35}13.23 & 0.37 & \cellcolor{red!35}0.53 \\
SparseGS~\cite{xiong2023sparsegs}
& 11.12 & 0.31 & 0.56 
& 11.86 & 0.36 & 0.57 
& 12.41 & 0.29 & 0.58 \\
FSGS~\cite{zhu2023FSGS}
& \cellcolor{yellow!35}12.89 & \cellcolor{yellow!35}0.46 & \cellcolor{yellow!35}0.47 
& \cellcolor{yellow!35}13.14 & \cellcolor{yellow!35}0.52 & 0.51 
& 13.22 & \cellcolor{yellow!35}0.40 & \cellcolor{yellow!35}0.54 \\
InstantSplat~\cite{fan2024instantsplat}
& 11.56 & 0.45 & 0.51 
& 11.42 & 0.48 & 0.53 
& 11.25 & 0.33 & 0.59 \\
GaMO~\cite{huang2025gamo}
& 11.25 & 0.35 & 0.60 
& 11.60 & 0.39 & 0.61 
& 12.88 & 0.34 & 0.60 \\
G4SPLAT~\cite{ni2026g4splat}
& 12.65 & \cellcolor{orange!35}0.51 & 0.51 
& \cellcolor{red!35}14.83 & \cellcolor{red!35}0.60 & \cellcolor{orange!35}0.47 
& \cellcolor{red!35}14.18 & \cellcolor{red!35}0.46 & 0.56 \\
\textbf{PanoPlane (Ours)}
& \cellcolor{red!35}\textbf{14.99} & \cellcolor{red!35}\textbf{0.53} & \cellcolor{red!35}\textbf{0.44} 
& \cellcolor{orange!35}\textbf{14.70} & \cellcolor{orange!35}\textbf{0.57} & \cellcolor{red!35}\textbf{0.44} 
& \cellcolor{orange!35}\textbf{13.74} & \cellcolor{orange!35}\textbf{0.41} & \cellcolor{orange!35}\textbf{0.53} \\
\midrule
\midrule
\multirow{2}{*}{\textbf{Method}} 
& \multicolumn{3}{c}{\textbf{ZsNo4HB9uLZ}} 
& \multicolumn{3}{c}{\textbf{RPmz2sHmrrY}} 
& \multicolumn{3}{c}{\textbf{2t7WUuJeko7}} \\
\cmidrule(lr){2-4} \cmidrule(lr){5-7} \cmidrule(lr){8-10}
& PSNR$\uparrow$ & SSIM$\uparrow$ & LPIPS$\downarrow$ 
& PSNR$\uparrow$ & SSIM$\uparrow$ & LPIPS$\downarrow$ 
& PSNR$\uparrow$ & SSIM$\uparrow$ & LPIPS$\downarrow$ \\
\midrule
3DGS~\cite{kerbl20233d}
& \cellcolor{orange!35}14.63 & \cellcolor{yellow!35}0.58 & \cellcolor{yellow!35}0.49 
& \cellcolor{yellow!35}15.35 & 0.51 & 0.50 
& \cellcolor{yellow!35}13.67 & 0.33 & \cellcolor{yellow!35}0.55 \\
SparseGS~\cite{xiong2023sparsegs}
& 13.20 & 0.46 & 0.53 
& 14.25 & 0.44 & 0.50 
& 13.31 & 0.32 & 0.56 \\
FSGS~\cite{zhu2023FSGS}
& \cellcolor{yellow!35}14.08 & \cellcolor{orange!35}0.65 & \cellcolor{orange!35}0.48 
& 15.11 & \cellcolor{yellow!35}0.55 & \cellcolor{yellow!35}0.49 
& 13.08 & 0.34 & 0.60 \\
InstantSplat~\cite{fan2024instantsplat}
& 12.54 & 0.55 & 0.51 
& 14.89 & 0.52 & 0.49 
& \cellcolor{orange!35}14.10 & \cellcolor{red!35}0.38 & \cellcolor{red!35}0.53 \\
GaMO~\cite{huang2025gamo}
& 11.36 & 0.52 & 0.61 
& 15.01 & 0.55 & 0.55 
& 13.32 & 0.31 & 0.59 \\
G4SPLAT~\cite{ni2026g4splat}
& 14.45 & 0.49 & 0.48 
& \cellcolor{orange!35}17.76 & \cellcolor{orange!35}0.63 & \cellcolor{orange!35}0.41 
& 13.53 & \cellcolor{orange!35}0.38 & 0.58 \\
\textbf{PanoPlane (Ours)}
& \cellcolor{red!35}\textbf{23.01} & \cellcolor{red!35}\textbf{0.66} & \cellcolor{red!35}\textbf{0.35} 
& \cellcolor{red!35}\textbf{22.09} & \cellcolor{red!35}\textbf{0.69} & \cellcolor{red!35}\textbf{0.32} 
& \cellcolor{red!35}\textbf{14.62} & \cellcolor{yellow!35}\textbf{0.35} & \cellcolor{orange!35}\textbf{0.53} \\
\bottomrule
\end{tabular}
}
\end{table*}

\section{Additional Qualitative Results}
\label{app:qual}

\begin{figure}[h]
  \centering
  \includegraphics[width=\textwidth]{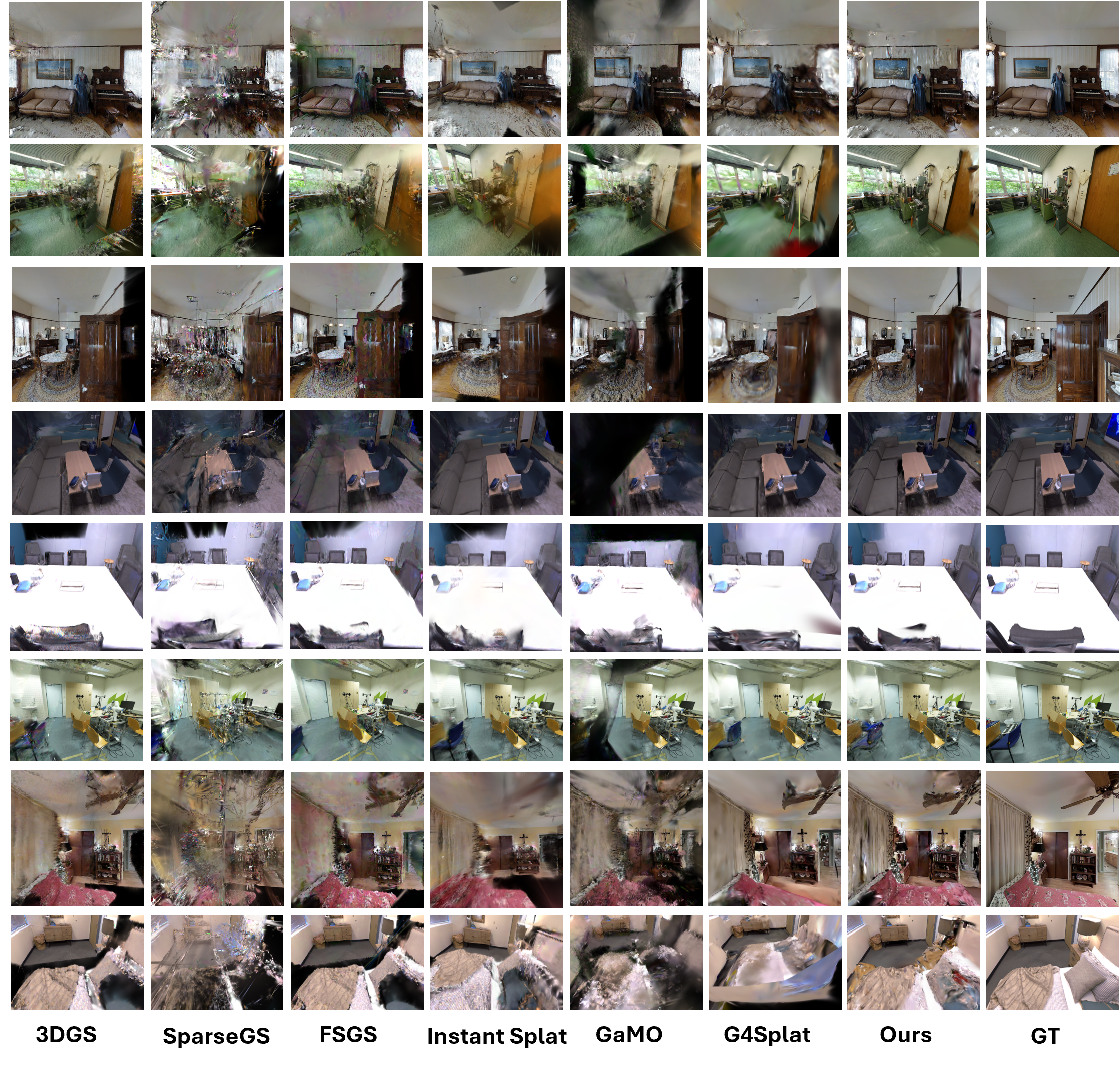}
  \caption{Additional Qualitative Results on Replica, Matterport3D and Scannet++}
\end{figure}

\begin{center}

  \includegraphics[width=\textwidth]{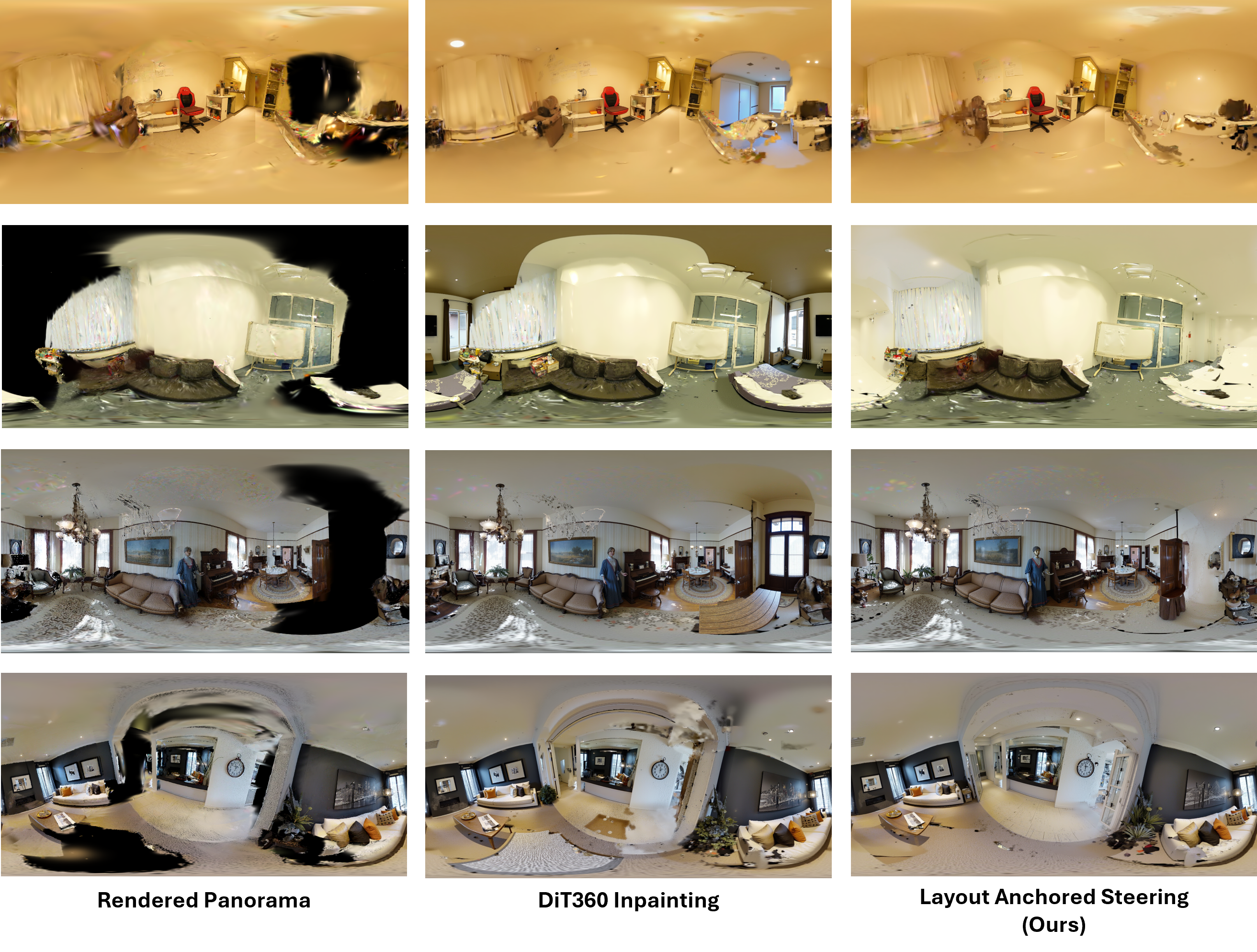}

    \captionof{figure}{\textbf{Additional qualitative results of layout-anchored steering:} Without steering (Naive), DiT360 inpaints unobserved regions according to its learned distribution, often hallucinating incorrect geometry such as receding corridors or warped surfaces. With our layout-anchored steering (Ours), the same regions are completed as flat extensions of the detected walls, ceilings, and floors, producing geometrically consistent completions that align with the observed room structure.}
  
\end{center}

\section{Use of Large Language Models}
\label{app:llm}
We employed a large language model for copy editing, including grammar checking, wording refinement, and minor improvements in style and clarity. This was done after we had completed the scientific content, methodology, analyses, and conclusions.

\end{document}